\pdfoutput=1
\documentclass[10pt,twocolumn,letterpaper]{article}

\usepackage[pagenumbers]{cvpr} % To force page numbers, \eg for an arXiv version

\usepackage{graphicx}
\usepackage{amsmath}
\usepackage{amssymb}
\usepackage{graphics}
\usepackage{booktabs}
\usepackage{multirow}
\usepackage{rotating}
\usepackage{url}
\usepackage{mathrsfs}
\usepackage[accsupp]{axessibility}
\usepackage{import}
\usepackage{bm}
\usepackage{float}
\usepackage[ruled,linesnumbered]{algorithm2e}
\usepackage{algpseudocode}

\usepackage[pagebackref,breaklinks,colorlinks]{hyperref}

\usepackage[capitalize]{cleveref}
\crefname{section}{Sec.}{Secs.}
\Crefname{section}{Section}{Sections}
\Crefname{table}{Table}{Tables}
\crefname{table}{Tab.}{Tabs.}

\def\cvprPaperID{5075} % *** Enter the CVPR Paper ID here
\def\confName{CVPR}
\def\confYear{2022}

\begin{document}

\title{Animal Kingdom: A Large and Diverse Dataset for Animal Behavior Understanding}

\author{
Xun Long Ng{\textsuperscript{‡}}\quad 
Kian Eng Ong{\textsuperscript{‡}}\quad
Qichen Zheng{\textsuperscript{‡}}\quad
Yun Ni{\textsuperscript{‡}}\quad 
Si Yong Yeo \quad
Jun Liu\thanks{Corresponding author.} \\
\normalsize Information Systems Technology and Design, 
Singapore University of Technology and Design, Singapore
\\
{\tt\scriptsize \{xunlong\_ng, kianeng\_ong\}@mymail.sutd.edu.sg} ~~~~~~~
{\tt\scriptsize \{qichen\_zheng, ni\_yun, siyong\_yeo, jun\_liu\}@sutd.edu.sg}
}

\maketitle

\def\thefootnote{{\textsuperscript{‡}}}\footnotetext{These authors contribute equally to this work.}

%%%%%%%%% ABSTRACT
\begin{abstract}
    \vspace{-0.3cm}
    
    Understanding animals' behaviors is significant for a wide range of applications. However, existing animal behavior datasets have limitations in multiple aspects, including limited numbers of animal classes, data samples and provided tasks, and also limited variations in environmental conditions and viewpoints. To address these limitations, we create a large and diverse dataset, Animal Kingdom, that provides multiple annotated tasks to enable a more thorough understanding of natural animal behaviors. The wild animal footages used in our dataset record different times of the day in extensive range of environments containing variations in backgrounds, viewpoints, illumination and weather conditions. More specifically, our dataset contains 50 hours of annotated videos to localize relevant animal behavior segments in long videos for the video grounding task, 30K video sequences for the fine-grained multi-label action recognition task, and 33K frames for the pose estimation task, which correspond to a diverse range of animals with 850 species across 6 major animal classes. Such a challenging and comprehensive dataset shall be able to facilitate the community to develop, adapt, and evaluate various types of advanced methods for animal behavior analysis. Moreover, we propose a Collaborative Action Recognition (CARe) model that learns general and specific features for action recognition with unseen new animals. This method achieves promising performance in our experiments.
    Our dataset can be found at \tt\footnotesize{\url{https://sutdcv.github.io/Animal-Kingdom}}.
    \end{abstract}

    \vspace{-0.5cm}
    
    \section{Introduction}

    A better understanding of how animals behave and move in the wild is not only a cornerstone of behavioral sciences \cite{graving2019deepposekit} but also crucial for a wide range of applications. Animal behavioral analysis plays a pivotal role in conservation efforts \cite{singh2020animal, ani11020485} and wildlife management \cite{nguyen2017animal}. Each year, the world spends at least \$75 billion \cite{mccarthy2012financial, biod2020global} in conservation efforts. Before researchers can better protect the wildlife, they often need to first monitor the animals. Researchers often monitor animals using video cameras, without having to attach physical markers that may alter their natural behaviors \cite{graving2019deepposekit}. By monitoring and analyzing the behaviors of the animals, they can gain deeper insights into their health and needs \cite{graving2019deepposekit}, and even detect movement-related injuries \cite{graving2019deepposekit}. In addition, detecting and analyzing changes in animal behaviors allow researchers to learn about new behaviors \cite{von2021big, ANDERSON201418}. Besides, animal behavioral analysis has also helped pharmaceutical scientists understand the effects of experimental interventions on behaviors, and neuroscientists learn more about brain activity across different temporal scales \cite{graving2019deepposekit, pereira2019fast, MATHIS20201, ANDERSON201418, karashchuk2021anipose}. These findings also help researchers better understand human diseases and develop drugs that are suitable for use in humans. Moreover, insights gained from analyzing animals' behaviors have also played an important role in aiding engineers in designing bio-inspired robots \cite{mu2020learning, zuffi2018lions} to efficiently perform specialised functions (\eg, lifesaving). Animators and game developers can also make use of the insights to realistically render animals in animations and games \cite{cao2019cross, zuffi2018lions}. In summary, the analysis of animal behaviors is not only significant for ecological protection, but also significant in a wide range of applications, and thus has received more and more research attention recently \cite{liwildlife, ani11020485, rahman2014fast, maaloy2019spatio, liu2020computer, yang2018feeding, yu2021apk, cao2019cross, mu2020learning, li2020deformation}.
    
    Animal action analysis is especially important in the understanding of the health and needs of animals. Given the diversity of animals’ actions, and how actions can occur and switch within split seconds, the use of video footages enables us to monitor multiple animals round the clock \cite{ANDERSON201418}. In addition, by localizing keypoints and employing pose estimation, we can analyze and identify changes in the animal's pose to better interpret their actions and behaviors. Before we analyze animal behaviors, we may need to first identify the frames of interest, since oftentimes a large part of the animal footages may not even contain any animal \cite{9522940, miao2021serengeti}. However, the vast volume of wildlife animal footages inherently make it both labour and time-intensive to identify animals and actions. In view of this, employing video grounding \cite{zhang2020span, mun2020LGI} will be effective to localize the animals and behaviors of interest in long video streaming.
    
    Previous studies \cite{von2021big, cao2019cross} show that a large and comprehensive dataset is important to develop robust deep learning models. However, many existing animal behavioral analysis datasets have limitations in multiple aspects: 
    \textbf{(1)} Relatively small dataset and scarcity of extensive and well-annotated labels \cite{von2021big}, which reduce the generalization ability and transferability of models \cite{von2021big}. In contrast, having a large and comprehensive dataset would be helpful to mitigate such issues. 
    \textbf{(2)} Limited number and limited diversity of animals \cite{ANDERSON201418, von2021big}, as many of the existing datasets are designed for and to study specific groups of animals (\eg, mammals only). However, there is a far greater diversity of animals in nature, and the understanding of wildlife should not be restricted to only a few specific classes. Furthermore, there exists a huge variety in shapes, sizes (\eg, different stages in life cycle), body patterns (\eg, camouflage), and number of limbs or appendages (\eg, wings) within and across animal classes. Therefore, having an extensive range and diverse representation of animals in a dataset would improve the generalization ability of the model for analysing behaviors of various animal classes \cite{von2021big}. 
    \textbf{(3)} Uniform environmental conditions (\eg, laboratory conditions or specific habitat) prevent the transferability of models to a different context (\eg, another habitat), because the models are often not able to generalize well when the new data differs from those previously used for training \cite{von2021big}. Moreover, animals in the wild are naturally found at various locations, in different environmental conditions, and at different time periods, which in turn will all affect the appearances and behaviors of the animals \cite{graving2019deepposekit}.  
    However, many of the existing datasets are confined to specific or uniform environmental set-ups. Therefore, it is essentially important to possess a sizable range of environmental conditions and backgrounds. 
    \textbf{(4)} Limited number of tasks and annotations for a more comprehensive analysis of animal behaviors. Most of the current datasets provide annotations of one task only, while different tasks can facilitate the understanding of animal behaviors in different aspects.
    
    Therefore, we create a diverse dataset, Animal Kingdom, for animal behavioral analysis from video grounding task that identifies and extracts the segments of relevant videos, to animal action recognition and pose estimation tasks to better understand animal behaviors. The natural behaviors of animals in the wild are generally dynamic, complex and noisy \cite{MATHIS20201, pereira2019fast, van2020deep}, and our diverse dataset is a good reflection of the realities in the wild. Besides constructing the Animal Kingdom dataset, in this paper, we also design a simple yet effective Collaborative Action Recognition (CARe) model for action recognition with unseen new types of animals.

    \section{Related Work}
    \subsection{Animal Behavioral Understanding Datasets}
    
    Analyzing poses and actions of animals in the wild allows researchers to objectively quantify natural behaviors \cite{graving2019deepposekit, ANDERSON201418, von2021big, MATHIS20201, pereira2019fast, robinson2014comparison} and uncover new behaviors \cite{von2021big, ANDERSON201418}. In the computer vision community, it has inspired lots of works for animal behavioral analysis \cite{liwildlife, ani11020485, rahman2014fast, maaloy2019spatio, liu2020computer, yang2018feeding, yu2021apk, cao2019cross, mu2020learning, li2020deformation, gupta2021dftnet, beyan2013detection, wang2020anomalous, rashid2017interspecies}. Despite the great demand for large unifying animal behavior datasets \cite{beery2021iwildcam, von2021big, ANDERSON201418}, most of the current datasets are still relatively small, disparate, animal and environment-specific. Moreover, these datasets are typically either mainly image-based or dedicated to animal classification tasks only (\eg, iNaturalist \cite{van2018inaturalist}, Animals with Attributes \cite{xian2020zeroshot}, Caltech-UCSD Birds \cite{WelinderEtal2010}, Florida Wildlife Animal Trap \cite{gagne2021florida}), or focused on one or few species only \cite{labuguen2021macaquepose, mathis2021pretraining, fang2021pose, WelinderEtal2010, neverova2021discovering}, or taken in specific environments (\eg, Snapshot Serengeti \cite{swanson2015snapshot}, Fish4Knowledge \cite{fish4knowledge}). We summarize some of the notable animal behavioral analysis datasets in \Cref{tab:comparison}.

    Many existing animal behavior datasets are tailored to specific environments and target only a few actions performed by specific animals. For example, the behaviors of mice \cite{VANDAM2020108536, geuther2021action}, worms \cite{javer2018open}, monkeys \cite{bala2020automated} and flies \cite{ravbar2019automatic, graving2019deepposekit} are often examined in the laboratory settings, sometimes to understand the effects of experimental interventions \cite{crispim2017my}. Behaviors of animals, such as sheep \cite{owoeye2018online}, cows \cite{liang2018benchmark}, and pigs \cite{liu2020computer} in livestock industry, and salmons \cite{maaloy2019spatio} in fishing and aquaculture industries, are most frequently studied. 
    
    Meanwhile, the understanding of wildlife behaviors for ecological understanding and environmental protection \cite{ani11020485} has garnered lots of interest of the community \cite{ani11020485, schindler2021identification, cao2019cross, yu2021apk, liwildlife}. However, most of them are still largely limited to a few or specific classes of animals. Animal Pose \cite{cao2019cross} is a smaller subset created from the publicly available VOC2011, which is targeted at only 5 mammal species and used solely for pose estimation. In contrast, by far one of the largest datasets available, AP-10K \cite{yu2021apk} focuses solely on pose estimation, and it is confined to only mammals. Likewise, the action recognition dataset in \cite{liwildlife} focuses solely on 7 action categories. Besides these, there are also some attempts to adopt synthetically generated animal data \cite{mu2020learning,shooter2021sydog} 
    for pose estimation on real-life images of animals. 
    
    In summary, most of these datasets focus specifically on a small number of action classes and types of animals, yet there exists far more animals out there in the animal kingdom. All of these also illustrate the evident challenges of obtaining considerably diverse and comprehensive set of animal data. In comparison to all the existing animal behavior datasets, our Animal Kingdom dataset demonstrates noteworthy advantages in terms of having a bigger dataset size for a number of tasks (\ie, video grounding, action recognition and pose estimation), with rich annotations of multi-label actions and a much vaster diversity of animal classes found in a wide range of diverse background scenes with different weather conditions, low light and night scenes, and taken at different viewpoints. Hence, our dataset provides a challenging and comprehensive benchmark for the community to develop and test different types of models for animal behavior analysis.

    \begin{table*}[t]
    \centering
    \caption{Comparison of our dataset with some of the previous animal behavioral analysis datasets. Our dataset contains more annotated data for various tasks, more diverse animal classes, multiple types of environmental and weather conditions, which shall provide a challenging and comprehensive benchmark for animal behavior understanding in images and videos.}
    \vspace{-0.3cm}
    \label{tab:comparison}
    \scalebox{0.58}{
    
    \begin{tabular}{|c||c||c|l|c|l|c|c||c||c|c||c|c||c||c|c|c|l|c|c|c|l|c||c|c|c|c|} 
    \hline
    \multirow{2}{*}{\begin{tabular}[c]{@{}c@{}}\\~\\\textbf{ Dataset}\end{tabular}} & \multirow{2}{*}{\begin{tabular}[c]{@{}c@{}}\textbf{Publicly}\\\textbf{available?}\end{tabular}} & \multicolumn{6}{c||}{\textbf{Diverse types of animals }} & \multirow{2}{*}{\begin{tabular}[c]{@{}c@{}}\textbf{No.}\\\textbf{of}\\\textbf{species }\end{tabular}} & \multicolumn{2}{c||}{\begin{tabular}[c]{@{}c@{}}\textbf{Task 1: Video}\\\textbf{Grounding~}\end{tabular}} & \multicolumn{2}{c||}{\begin{tabular}[c]{@{}c@{}}\textbf{Task 2: Action}\\\textbf{Recognition}\\\end{tabular}} & \begin{tabular}[c]{@{}c@{}}\textbf{Task 3: Pose}\\\textbf{Estimation}\\\end{tabular} & \multicolumn{9}{c||}{\textbf{Types of scene }} & \multicolumn{4}{c|}{\textbf{Weather }} \\ 
    \cline{3-8}\cline{10-27}
     &  & \begin{sideways}Mammals\end{sideways} & \begin{sideways}Reptiles\end{sideways} & \begin{sideways}Birds\end{sideways} & \begin{sideways}Amphibians\end{sideways} & \begin{sideways}Fishes\end{sideways} & \begin{sideways}Insects\end{sideways} &  & \begin{sideways}\begin{tabular}[c]{@{}c@{}}No. of annotated\\long videos\end{tabular}\end{sideways} & \begin{sideways}No. of statements\end{sideways} & \begin{sideways}\begin{tabular}[c]{@{}c@{}}No. of annotated\\video clips\end{tabular}\end{sideways} & \begin{sideways}\begin{tabular}[c]{@{}c@{}}No. of annotated\\action classes\end{tabular}\end{sideways} & \begin{sideways}\begin{tabular}[c]{@{}c@{}}No. of labelled\\images\end{tabular}\end{sideways} & \begin{sideways}Night scene\end{sideways} & \begin{sideways}Low light\end{sideways} & \begin{sideways}Complex background\end{sideways} & \begin{sideways}Mountain\end{sideways} & \begin{sideways}Forest\end{sideways} & \begin{sideways}Grassland\end{sideways} & \begin{sideways}Desert\end{sideways} & \begin{sideways}Ocean\end{sideways} & \begin{sideways}Underwater\end{sideways} & \begin{sideways}Windy\end{sideways} & \begin{sideways}Foggy\end{sideways} & \begin{sideways}Rain\end{sideways} & \begin{sideways}Snow\end{sideways} \\ 
    \hline
    Broiler Chicken \cite{fang2021pose} & $\times$ & $\times$ & $\times$ & $\checkmark$ & $\times$ & $\times$ & $\times$ & NA & $\times$ & $\times$ & NA & 6 & 556 & NA & NA & NA & $\times$ & $\times$ & $\times$ & $\times$ & $\times$ & $\times$ & NA & NA & NA & NA \\ 
    \hline
    Fish Action \cite{rahman2014fast} & $\times$ & $\times$ & $\times$ & $\times$ & $\times$ & $\checkmark$ & $\times$ & NA & $\times$ & $\times$ & 95 & 5 & $\times$ & $\times$ & $\checkmark$ & $\checkmark$ & $\times$ & $\times$ & $\times$ & $\times$ & $\checkmark$ & $\checkmark$ & $\times$ & $\times$ & $\times$ & $\times$ \\ 
    \hline
    Salmon Feeding \cite{maaloy2019spatio} & $\times$ & $\times$ & $\times$ & $\times$ & $\times$ & $\checkmark$ & $\times$ & 1 & $\times$ & $\times$ & 76 & 2 & $\times$ & $\times$ & $\checkmark$ & $\checkmark$ & $\times$ & $\times$ & $\times$ & $\times$ & $\checkmark$ & $\checkmark$ & $\times$ & $\times$ & $\checkmark$ & $\times$ \\ 
    \hline
    Wild Felines \cite{ani11020485} & $\times$ & $\checkmark$ & $\times$ & $\times$ & $\times$ & $\times$ & $\times$ & 3 & $\times$ & $\times$ & 2,700 & 3 & $\times$ & $\checkmark$ & $\checkmark$ & $\checkmark$ & $\times$ & $\checkmark$ & $\checkmark$ & $\times$ & $\times$ & $\times$ & NA & NA & NA & NA \\ 
    \hline
    Pig Tail-biting \cite{liu2020computer} & $\times$ & $\checkmark$ & $\times$ & $\times$ & $\times$ & $\times$ & $\times$ & 1 & $\times$ & $\times$ & 4,396 & 2 & $\times$ & $\times$ & $\checkmark$ & $\times$ & $\times$ & $\times$ & $\times$ & $\times$ & $\times$ & $\times$ & $\times$ & $\times$ & $\times$ & $\times$ \\ 
    \hline
    Wildlife Action \cite{liwildlife} & $\times$ & $\checkmark$ & $\checkmark$ & $\checkmark$ & \textbf{\textbf{$\checkmark$}} & $\checkmark$ & $\checkmark$ & 106 & $\times$ & $\times$ & 10,600 & 7 & $\times$ & $\checkmark$ & $\checkmark$ & $\checkmark$ & $\times$ & $\checkmark$ & $\checkmark$ & $\times$ & $\checkmark$ & $\checkmark$ & NA & NA & NA & NA \\ 
    \hline
    Animal Pose \cite{cao2019cross} & $\checkmark$ & $\checkmark$ & $\times$ & $\times$ & $\times$ & $\times$ & $\times$ & 5 & $\times$ & $\times$ & $\times$ & $\times$ & 4,666 & $\checkmark$ & $\checkmark$ & $\checkmark$ & $\checkmark$ & $\checkmark$ & $\checkmark$ & $\times$ & $\times$ & $\times$ & $\times$ & $\checkmark$ & $\times$ & $\checkmark$ \\ 
    \hline
    Horse-30 \cite{mathis2021pretraining} & $\checkmark$ & $\checkmark$ & $\times$ & $\times$ & $\times$ & $\times$ & $\times$ & 3 & $\times$ & $\times$ & $\times$ & $\times$ & 8,144 & $\times$ & $\times$ & $\checkmark$ & $\times$ & $\times$ & $\checkmark$ & $\times$ & $\times$ & $\times$ & $\times$ & $\times$ & $\times$ & $\times$ \\ 
    \hline
    AP-10K \cite{yu2021apk} & $\checkmark$ & $\checkmark$ & $\times$ & $\times$ & $\times$ & $\times$ & $\times$ & 54 & $\times$ & $\times$ & $\times$ & $\times$ & 10,015 & $\checkmark$ & $\checkmark$ & $\checkmark$ & $\checkmark$ & $\checkmark$ & $\checkmark$ & $\times$ & $\checkmark$ & $\checkmark$ & $\times$ & $\checkmark$ & $\times$ & $\checkmark$ \\ 
    \hline
    Macaque Pose \cite{labuguen2021macaquepose} & $\checkmark$ & $\checkmark$ & $\times$ & $\times$ & $\times$ & $\times$ & $\times$ & NA & $\times$ & $\times$ & $\times$ & $\times$ & 13,083 & $\checkmark$ & $\checkmark$ & $\checkmark$ & $\checkmark$ & $\checkmark$ & $\times$ & $\times$ & $\times$ & $\times$ & $\times$ & $\checkmark$ & $\times$ & $\checkmark$ \\ 
    \hline
    Dogs \cite{barnard2016quick} & $\checkmark$ & $\checkmark$ & $\times$ & $\times$ & $\times$ & $\times$ & $\times$ & 1 & $\times$ & $\times$ & 13 & 4 & 2,200 & $\times$ & $\checkmark$ & $\times$ & $\times$ & $\times$ & $\times$ & $\times$ & $\times$ & $\times$ & $\times$ & $\times$ & $\times$ & $\times$ \\ 
    \hline
    \begin{tabular}[c]{@{}c@{}}\textbf{Animal Kingdom}\\\textbf{(Ours) }\end{tabular} & $\checkmark$ & \textbf{$\checkmark$} & \textbf{\textbf{$\checkmark$}} & \textbf{$\checkmark$} & \textbf{\textbf{$\checkmark$}} & \textbf{$\checkmark$} & \textbf{$\checkmark$} & \textbf{850} & \begin{tabular}[c]{@{}c@{}}\textbf{4,301}\\\textbf{(50h)}\end{tabular} & \textbf{\textbf{18,744}} & \begin{tabular}[c]{@{}c@{}}\textbf{30,100}\\\textbf{(50h)}\end{tabular} & \textbf{140 } & \textbf{33,099} & \textbf{$\checkmark$} & \textbf{$\checkmark$} & \textbf{$\checkmark$} & $\checkmark$ & \textbf{$\checkmark$} & \textbf{$\checkmark$} & \textbf{$\checkmark$} & \textbf{$\checkmark$} & \textbf{$\checkmark$} & \textbf{$\checkmark$} & \textbf{$\checkmark$} & \textbf{$\checkmark$} & \textbf{$\checkmark$} \\
    \hline
    \end{tabular}
    }

    \vspace{-0.5cm}
    \end{table*}

    \subsection{Animal Behavioral Analysis Methods}
    Some animal pose estimation models have been proposed \cite{mathis2018deeplabcut, pereira2019fast, graving2019deepposekit, karashchuk2021anipose}. Various body parts of the animals are segmented and used for pose normalization in \cite{tang2020revisiting}. Some works use synthetic images to learn poses of real animals \cite{li2020deformation, li2021synthetic,shooter2021sydog}. 3D models \cite{mu2020learning, zuffi20173d} are even used in this endeavour. Another approach is to leverage knowledge from synthetic to real animals \cite{li2021synthetic}, or from both human and animal poses \cite{cao2019cross} to extend pose knowledge to other animals. 
    
    Current animal action recognition models are mainly adapted from existing human action recognition methods. Some of the popular ones are CNN \cite{bohnslav2021deepethogram} with LSTM \cite{owoeye2018online, liu2020computer}, Mask R-CNN \cite{ani11020485, schindler2021identification} with VGG\cite{ani11020485}, I3D \cite{liwildlife}, (2+1)D ResNet \cite{schindler2021identification}, and SlowFast \cite{schindler2021identification}. Different from existing animal action recognition methods, in this work, considering 
    the diversity of animals and the fact that the same class of action (e.g., \textit{eating}) can look similar yet different when performed by different types of animals, we develop a simple yet effective Collaborative Action Recognition (CARe) model that extracts and utilizes both general and specific features to effectively identify the actions of unseen new types of animals.   
    
    \vspace{-0.2cm}
    \section{The Proposed Animal Kingdom Dataset}
    \subsection{Dataset Description}
    Our Animal Kingdom dataset contains 50 hours of long videos for video grounding, 50 hours of video clips for action recognition, and 33K annotated frames for pose estimation. In addition, our dataset possesses significant properties as listed below.

\begin{figure}[t]
    
        \centering
    
        \includegraphics[width=0.48\textwidth]{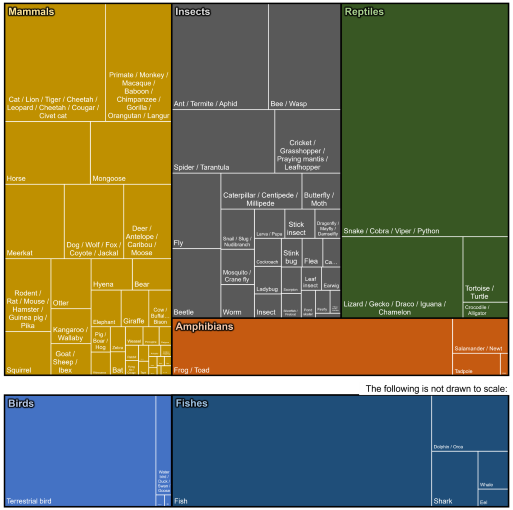}
        \vspace{-0.6cm}
        \caption{Distribution of clips of over 850 species of animals in 6 major animal classes, classified based on their appearance, number of limbs and how they move, and further divided into sub-classes.}
        \label{fig:dist_animal}
        \vspace{-0.3cm}
    \end{figure}

    \textbf{Diverse types of animals and actions.}
    Having a large diversity of animals and actions is useful to understand animal behaviors in the wild, and important for training various types of models. Our dataset identifies 140 classes of actions exhibited by a wide range of terrestrial and aquatic animals, which make up to 850 unique species that span over 6 key animal classes (\eg, reptiles, birds, and fishes as seen in \cref{fig:dist_animal}). Even for the same animal, the behaviors differ in different scenarios (\eg, amphibious animals like frogs which live and move differently in water and on land). Such is the intriguing lives of animals which add dimensions of complexities to action recognition in our dataset, and also influence the postures and movements of the animals, thus giving rise to a variety of animal poses. On the other hand, some animals having the same outward appearance live in totally different habitats and thus have different forms of movements (\eg, a terrestrial spider that crawls on land but a water spider that swims in water). All of these examples in our dataset illustrate the complex yet beautiful diversity of wildlife, and highlight the benefit of containing a diverse and good range of animal classes to encompass various actions and behaviors that occur naturally in the wild.

    \begin{figure}[t] 
    
        \centering
        \includegraphics[width=1\columnwidth]{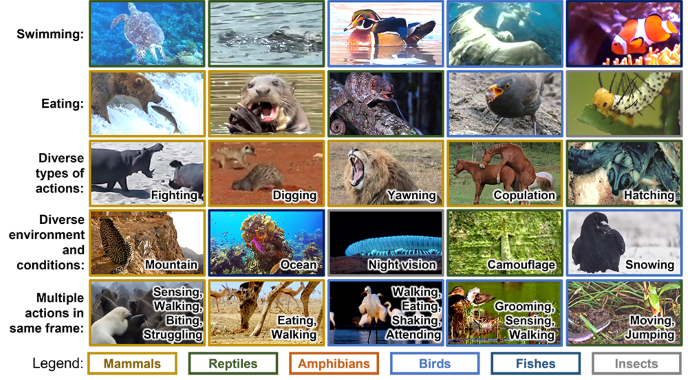}
        \vspace{-0.6cm}
        \caption{Examples of actions across various animal classes in our dataset. Rows 1 and 2 show how the same set of actions differs across various animal classes. Row 3 shows examples of various actions in our dataset. Row 4 shows sample features of our dataset, ranging from diverse environments to varying illumination and weather conditions. Row 5 shows multiple actions performed by multiple animals in the same frame. 
        }
     
        \label{fig:eg_ar}
        \vspace{-0.5cm}
    \end{figure}
    
    \textbf{Diverse environmental conditions.}
    The various environmental and weather conditions enable better understanding of animal behaviors in natural ecosystems. In the wild, animals are found in a variety of environments (\eg, grassland, forests, rivers, oceans) and environmental conditions (\eg, raining, snowing). The animal footages in our dataset are captured at different times of day, under different illumination conditions (\eg low light, sunset, backlight, different underwater depths), sometimes leading to shadow cast and low foreground-background contrast. This challenge is compounded when some animals (\eg, crows or ravens in \cref{fig:eg_ar}) lack distinctive colours for various body parts, while others (\eg, tree lizards in \cref{fig:eg_ar}) camouflage very well, blending seamlessly into their environment \cite{risse2017visual}. Some animals (\eg, chameleons in \cref{fig:eg_ar}) may even change their appearance rapidly in response to the environment. Animals move around in habitats with complex compositions and dynamic background with other animals in close proximity, or plants swaying in the wind \cite{risse2017visual}. All of these are found in our dataset and present the significant yet practical challenges of animal behavioral analysis in the wild.

    \textbf{Varied viewpoints and various types of footages.}
    Obtaining real footages documenting how animals live and behave in the wild is of great significance to ecological studies and conservation efforts \cite{singh2020animal}. The animal footages, which were captured by nature enthusiasts and professionals who used different equipment (\eg, night vision), come in various forms (\eg, documentaries, animal trap videos, and personal camera recordings), and were taken from various vantage points (\eg, birds' eye view, bottom up, underwater). This brings diversities of captured viewpoints in our dataset.
    
    \textbf{Various compositions of scenes with fine-grained multi-label actions.} Our dataset encapsulates the lives and actions of wildlife, with clips depicting the fine-grained action(s) of one animal to multiple types of animals in a single frame. This leads to the creation of a multi-label dataset comprising fine-grained actions to aptly describe each unique action of animal(s) in the clip. However, this also inevitably leads to a long-tailed distribution of actions, which poses a significant challenge for accurate models, and deserves attention to deepen research into the building of more robust models for the practical world. \par

    \subsection{Dataset Tasks and Annotations}
    
    Similar to the experience detailed in \cite{von2021big}, our diverse dataset for video grounding, action recognition and pose estimation was meticulously assembled together through the collective effort of 23 individuals (including biology experts). We manually identified and provided framewise annotations of both animal and action descriptions for 50 hours of videos that were collected from YouTube videos. A total of 3 rounds of quality checks were performed to uphold the annotation quality. Please refer to the Supplementary and our website for more details of our dataset. 
    
    \textbf{Action recognition with multi-label fine-grained actions.} In action recognition, the model takes in an input video clip, and outputs the action labels for it. Actions of animals can occur within split seconds (\eg, jumping), and up to minutes (\eg, sexual display) for more complex behavioral patterns. Our diverse dataset contains 50 hours of video clips, comprising over 850 animal species and 140 fine-grained action classes that were taken from a list used by ethologists \cite{mouse_ethogram, stafford2011inferential, rose2021conducting, NC3RS_ethogram}. The collection of actions and behaviors (see Supplementary) encompasses life events (\eg, molting), to daily activities (\eg, feeding), to social interactions (\eg, playing). The video clips last an average of 6 seconds, and range from 1 second to 117 seconds.

    \label{subsection:long-tail}
    The complexities brought by the natural world, present a set of practical challenges in analysing animal actions in the wild. 
    \textbf{(1)} There exists a large intra and inter-class variation of actions in our dataset. As seen in the first two rows in \cref{fig:eg_ar}, the same type of action (\eg, eating) can look differently when performed by different types of animals. Hence, we need to design robust methods to recognize the actions of various animals. 
    \textbf{(2)} In nature, an animal can perform more than one action at the same time. Besides, there can be multiple animals in the same frame, with different animals performing different actions. All of these lead to the multi-label actions, and can also result in a long-tailed distribution of actions. 
    \textbf{(3)} As the videos of animals were captured in the wild, some actions (\eg, molting) occur less frequently than others (\eg, eating) in nature. This inevitably leads to an uneven long-tailed distribution of actions, thus encouraging the development of advanced strategies for handling long-tailed distributions.
    
    When constructing our animal action recognition dataset,
    we follow the work of \cite{zhang2021videolt} and divide the distribution into three different segments based on the number of samples in each action class. Specifically, we group all the 140 action classes in our dataset into the head segment (17 action classes that have more than 500 samples each), the middle segment (29 action classes that have 100 to 500 samples each), and the tail segment (94 action classes that have fewer than 100 samples each). The head segment contains the more frequently occurring actions in nature (\eg, sensing and eating). The middle segment consists of actions that are commonly observed in nature (\eg, climbing, grooming, and digging), while the tail segment is made up of actions that are less frequently observed in nature (\eg, molting and doing somersault).  
    
\begin{figure}[h]
        \centering
    
        \includegraphics[width=1\columnwidth]{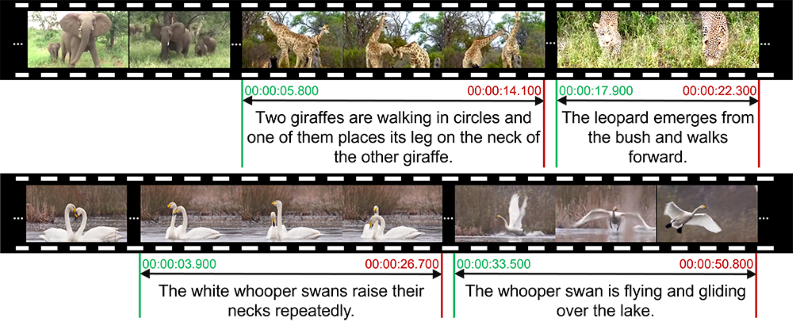}
        \vspace{-0.8cm}
        \caption{Samples of the video grounding task. Given the language description, we need to detect the corresponding time segment.
        }
        \label{fig:eg_vg}
        \vspace{-0.5cm}
    \end{figure}

    \textbf{Video grounding based on descriptions.}
    In video grounding, an input query sentence, describing the scene with the animals and behaviors of interest, is provided to the model. Then the model needs to output the relevant temporal segment with the start and end timing (\cref{fig:eg_vg}), much like a video clip search engine. Video grounding plays a pivotal role in improving productivity for ecological research, because a large part of the ecological surveillance videos may not contain the footage of the animals and behaviors of interest \cite{9522940, miao2021serengeti}, while video grounding provides a convenient way for users to search for the relevant time segment in long videos by describing the scenarios of interest.

    The key challenge of video grounding comes in the form of identifying the behaviors and delineating the animals from the complex background to return relevant results based on the language descriptions. The video grounding task in our dataset contains 50 hours of videos (a total of 4301 long video sequences) with 18,744 annotated sentences. Each video sequence contains 3-5 sentences.  
    
\begin{figure}[b]
    \vspace{-0.2cm}
      \centering
      \includegraphics[width=\linewidth] {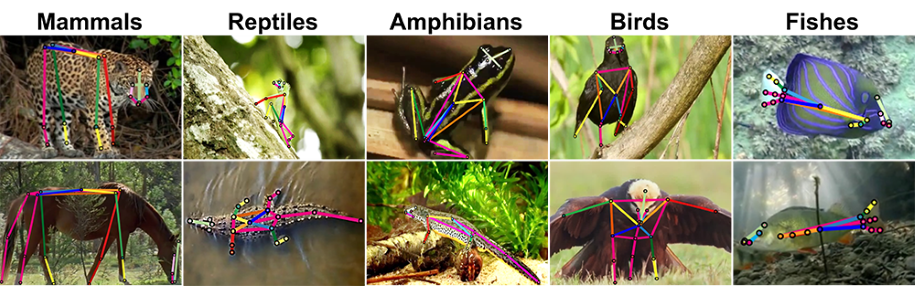}
    
      \vspace{-0.2cm}
        \caption{Examples of animal poses in our dataset}
        \label{fig:eg_pose}
    
    \end{figure}

    \newpage
    \textbf{Pose estimation for inferring animal keypoints.}
    In pose estimation, the model takes in an input image of the animal, and predicts the locations of its joints. As animals can have vastly different anatomical structures across classes, this poses a challenge for pose estimation. As such, we organize the images according to five main animal classes: mammals, reptiles, amphibians, birds, and fishes, to construct differentiated understanding of the poses for each animal class. We institute a common set of keypoints across the five animal classes (\cref{fig:eg_pose}). These keypoints correspond to the equivalent of human poses, and are also in line with the recommendations in \cite{von2021big}. We define a total of 23 keypoints: 1 head, 2 eyes, 4 parts of mouth, 2 shoulders, 2 elbows, 2 wrists, 1 mid torso, 2 hips, 2 knees, 2 ankles, and 3 parts of the tail. For the upper limbs of birds (\ie, wings) and fishes (\ie, fins), their shoulders, elbows, and wrists are defined in accordance to how their upper limbs move. Similarly, for the `lower' limbs of fishes which do not have legs, their hips, knees, and ankles are annotated along the edge of their tail section, because their tails control their swimming movements \cite{borazjani2013fish}. \par
    
    In summary, our Animal Kingdom dataset possesses several significant properties for various types of animal behavior analysis tasks. In addition, the large intra-class variations, multi-label actions, and long-tailed distributions, present practical challenges to understanding animal behaviors in the wild. Hence, our dataset provides a challenging and diverse benchmark to facilitate the research community to develop, adapt, and evaluate various types of advanced methods for animal behavioral analysis.

    \section{\textbf{The Proposed CARe Model}}

    When constructing our dataset, we observe that the same class of action can look similar yet different when performed by different types of animals. For example, the action ``eating" shares similarities yet still differs between chameleons and otters, as shown in \cref{fig:eg_ar}. Meanwhile, training network models to analyze and recognize actions of various types of animals can be an interesting problem. However, even if we can collect a large number of video action samples for a set of animals, it can still be difficult to exhaustively 
    collect and annotate lots of samples for various types of species for network training, considering that there are more than 7 million species in the world \cite{mora2011many}. 
    
    Nevertheless, since the same action class of different animals can share some common (generalizable) characteristics, it becomes possible to train a general network based on video samples of the annotated animal set, and then apply this network for recognizing the same set of actions of even unseen new animals (\ie, animal types that are not included during model training, but are of interest later).
    In this way, well generalizable features shared across different animals can be extracted by the network for action recognition. 
    However, extracting general features only while ignoring the specific features of actions of different animal types can limit the action recognition performance, because the specific features, which though may be unique to a certain animal type only, can still convey discriminative information for recognizing the actions of this animal type.

    \begin{figure}[t]
      \centering
    
            \includegraphics[width=1\linewidth]{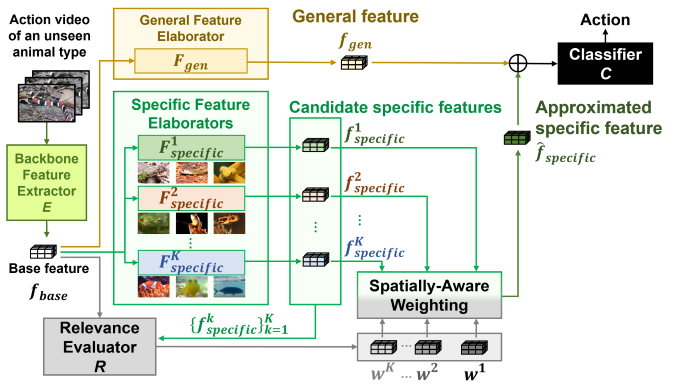}
            \vspace{-0.8cm}
        \caption{Architecture of our Collaborative Action Recognition (CARe) model. During testing, given an video action sample of an unseen type of animal, the approximated specific feature will be obtained by combining the outputs of the $K$ existing specific feature elaborators in a spatially-aware weighting manner. Then the approximated specific feature will be fused with the general feature, and finally fed to the classifier $C$ to predict the action class.
        }
        \label{fig:architecture}
        \vspace{-0.5cm}
    \end{figure}
    
    To address this issue, we propose a simple yet effective Collaborative Action Recognition model (CARe), which is able to recognize the same set of actions even for unseen types of animals, by leveraging both the general and specific features, as shown in \cref{fig:architecture}. Assuming that we have the data of $K$ types of observed animals for training, then the training data can be denoted as $\textbf{D}=\{D_k\}_{k=1}^{K}$, where $D_k=\{(x^k_n,y^k_n)\}_{n=1}^{N_k}$. Here $x^k_n$ represents a video action sample, $y^k_n$ represents its ground-truth action label, and $N_k$ is the number of samples for the $k$-th observed animal type. 
    
    Our basic model is composed of the backbone feature extractor $E$, as well as the relatively lightweight general feature elaborator ${F_{gen}}$ (shared by all animal types) and specific feature elaborators $\{F^k_{specific}\}_{k=1}^{K}$ (each specific for a certain type of animal), and the classifier $C$. The backbone feature extractor $E$ (an I3D model) is used to extract base feature  
    $f_{base}=E(x)$ from each input video sample $x$,
    which is then fed to the general feature elaborator ${F_{gen}}$ to get the general feature $f_{gen}={F_{gen}}(f_{base})$, and a specific feature elaborator $F^k_{specific}$ to get the specific feature $f^k_{specific} = F^k_{specific}(f_{base})$.
    
    Such a base model can be trained on the training set $\textbf{D}$ as follows:
    
    \vspace{-0.6cm}
    \begin{equation}
    \label{eq_loss}
    \begin{aligned}
        \hat{y}_n^k &= C\Big(F_{gen}\big(E(x_n^k)\big), F^k_{specific}\big(E(x_n^k)\big)\Big) \\
        \ell&=\ell_\textrm{CE}(\hat{y}_n^k, y_n^k)
        \end{aligned}
    \end{equation}
    \vspace{-0.4cm}
    
    This indicates, for each training sample $x_n^k$, its general and specific features are concatenated and finally fed to the classifier $C$ for predicting the action label $\hat{y}_n^k$, and the training of this base model is performed under the supervision of the cross-entropy loss $\ell_\textrm{CE}$. Via this scheme, during training, action samples from all the $K$ types of training animals will pass through the general feature elaborator ($F_{gen}$), while only the samples from the $k$-th animal type will pass through the specific elaborator $F^k_{specific}$. Thus, the general feature elaborator (that is well trained on a large amount of data of various (\ie, $K$) types of animals) is able to learn to extract well generalizable features. Meanwhile, each specific elaborator $F^k_{specific}$ (that is trained on the data of a specific type of animal) will learn to extract features that tend to be more specific for the corresponding type of animal. Thus, these two sets of features, which are complementary, can be fused for action recognition.

    Though the aforementioned basic model can generate general and specific features for a seen animal type, it still does not address the scenario for handling unseen animal types, since we do not have specific feature elaborators for animal types beyond these $K$ trained types.

    Considering that different animals can share different levels of similarities in their appearances and behaviors (\eg, cheetahs share many more similarities with tigers, compared to other animals like snakes), when handling an input video of an unseen animal type, to obtain its specific feature, a possible solution is to leverage existing elaborators based on different levels of similarities, to mine and approximate the specific feature for the unseen animal.
    
    To this end, we design a relevance evaluator $R$ that is a sub-network for assessing the similarities of the action sample ($x$) of the unseen animal type w.r.t all the $K$ observed animal types.
    To achieve this, we first input $x$ to the backbone feature extractor $E$ to get the base feature ($f_{base}$), which is then fed to all the $K$ specific elaborators, yielding a set of intermediate features $\{f^k_{specific}\}_{k=1}^K$.
    Then we estimate the similarities as: $\textbf{w}=R\big(f_{base}, \{f^k_{specific}\}_{k=1}^K \big)$, where $\textbf{w} = \{w^k\}_{k=1}^K$ represents the set of similarity (relevance) scores of this sample w.r.t the $K$ observed animal types.
    Thus, we can then approximate the specific feature of this unseen animal type's action sample ($x$), via the collaboration of the existing $K$ specific elaborators as:
    
    \vspace{-0.3cm}
    \begin{equation} \label{eq_moe}
    {\hat f}_{specific} = \frac{1}{K}\sum_{k=1}^{K} w^k   \cdot f^k_{specific}
    \end{equation}
    
    \vspace{-3mm}
    
    Moreover, considering an animal can have different similarity degrees w.r.t another type of animal at different spatial positions (different body parts), $w^k$ can also be implemented as a tensor (\ie, $w^k \in \mathbb{R}^{h \times w}$) instead of a scalar, such that the approximated specific feature of the unseen animal type is produced based on its different similarities to different observed animal types and also different spatial positions on the feature maps (with size $h \times w$). In this way, the $K$ feature elaborators collaborate with one another in a spatially-aware manner to yield optimal approximation. After specific features have been approximated, the general and approximated specific features are fused and fed to the classifier $C$ to predict the action class, as shown in \cref{fig:architecture}.

    As mentioned before, our basic model can be trained on the training set ($\textbf{D}$) of the observed animals. However, the relevance evaluator $R$ needs to be generalizable to unseen new animals,  
    which makes the training of $R$ challenging. 
    Inspired by meta learning \cite{dai2021generalizable, li2018learning,finn2017model, guo2020learning}, which is a ``learning to learn'' technique that can learn to generalize by splitting the original training set to meta-train and meta-test sets to mimic target testing scenarios during model training, we here apply meta-learning to train our $R$ for improving its generalization ability to unseen animal types.  
    Specifically, following the mimicking scheme of meta learning, 
    we split the training set containing the $K$ types of animals into a virtual meta-train set (with $K-1$ seen types of animals), and a virtual meta-test set (with one ``unseen'' type of animal). Based on these sets, we then take advantage of the domain generalization meta-learning (MLDG) algorithm \cite{li2018learning} that can effectively optimize our module $R$.
    
    By alternately training our base model using \cref{eq_loss} and training $R$ via MLDG, our CARe model obtains generalization ability, and can produce reliable general and approximated specific features that are useful for recognition of the same set of actions of unseen animals.  
    More implementation details of CARe can be found in the Supplementary.

    \section{Experiments}
    We evaluate state-of-the-art methods on our dataset: I3D \cite{carreira2017quo}, X3D \cite{feichtenhofer2020x3d}, and SlowFast \cite{feichtenhofer2019slowfast} for action recognition; VSLNet \cite{zhang2020span} and LGI \cite{mun2020LGI} for video grounding; and HRNet \cite{SunXLW19} and HRNet-DARK\cite{Zhang_2020_CVPR} for pose estimation. We use their original codes and adapt them for our Animal Kingdom dataset.
    
    \subsection{Action Recognition Results}
    
    For our first type of experiments, we perform multi-label action recognition. We first use Multi-Label Binarizer \cite{scikit-learn} on our multi-label animal action recognition dataset, which contains all the 30.1K video clips, to stratify the action classes and split both the video clips and each action class into 80\% train and 20\% test. Following \cite{sigurdsson2016hollywood}, we adopt the mean Average Precision (mAP) as our evaluation metric. 
    As mentioned in \cref{subsection:long-tail}, our dataset contains a long-tailed distribution of multi-label actions and we follow \cite{zhang2021videolt} to identify the head, middle and tail segments. Thus, we evaluate the overall mAP and also segment-specific mAP of action recognition. Note that as our dataset has long-tailed distribution, we also evaluate methods for handling long-tailed distribution \cite{lin2017focal, cao2019learning, tan2020equalization} on our dataset, and apply these methods on action recognition methods. Specifically, we test the cost-sensitive learning  (Focal Loss \cite{lin2017focal}) and re-weighting methods (LDAM-DRW \cite{cao2019learning} and EQL \cite{tan2020equalization}), in conjunction with different action recognition methods (I3D \cite{carreira2017quo}, X3D \cite{feichtenhofer2020x3d}, and SlowFast \cite{feichtenhofer2019slowfast}). As shown in \Cref{table:ar}, action recognition methods with approaches for handling long-tailed distribution (\eg, Focal Loss) achieve higher mAP for the head, middle and tail segments, compared to the baselines. This shows the possibility and brings opportunities for the community to explore various approaches to handle challenges including long-tailed and multi-label problems.

    \begin{table}[t]
      \centering
      \caption{Results of action recognition}
      \label{table:ar}
      \vspace{-0.3cm}
      \scalebox{0.69}{
      \begin{tabular}{clclclclcl}
        \toprule
        & \multicolumn{4}{c}{mAP}\\
        \cmidrule(lr){2-5}
        Method & overall & head & middle & tail\\
        \midrule
        \multicolumn{5}{c}{Baseline (Cross Entropy Loss)}\\
        \cmidrule(lr){1-5}
        I3D \cite{carreira2017quo}        & 16.48   &  46.39  &  20.68  & 12.28 \\
        SlowFast \cite{feichtenhofer2019slowfast}    & 20.46   & 54.52   &  27.68  & 15.07 \\
        X3D \cite{feichtenhofer2020x3d}       & 25.25   &  60.33  & 36.19   & 18.83 \\
        \cmidrule(lr){1-5}
        \multicolumn{5}{c}{Focal Loss \cite{lin2017focal}}\\
        \cmidrule(lr){1-5}
        I3D \cite{carreira2017quo}        &  26.49  & 64.72   &  40.18  & 19.07 \\
        SlowFast \cite{feichtenhofer2019slowfast}    & 24.74   & 60.72   & 34.59   & 18.51 \\
        X3D \cite{feichtenhofer2020x3d}       & 28.85   & 64.44   & 39.72   & 22.41 \\
        \cmidrule(lr){1-5}

        \multicolumn{5}{c}{LDAM-DRW \cite{cao2019learning}}\\
        \cmidrule(lr){1-5}
        I3D \cite{carreira2017quo}        &  22.40  &  53.26  &  27.73  & 17.82 \\
        SlowFast \cite{feichtenhofer2019slowfast}    &  22.65 &  50.02  & 29.23   & 17.61 \\
        X3D \cite{feichtenhofer2020x3d}       &  30.54  & 62.46   & 39.48   & 24.96 \\
      \cmidrule(lr){1-5}
       
      \multicolumn{5}{c}{EQL \cite{tan2020equalization}}\\
        \cmidrule(lr){1-5}
        I3D \cite{carreira2017quo}        &  24.85  &  60.63  &  35.36  &  18.47\\
        SlowFast \cite{feichtenhofer2019slowfast}    &  24.41 &  59.70  & 34.99   & 18.07 \\
        X3D \cite{feichtenhofer2020x3d}    & 30.55  &  63.33  &  38.62  &  25.09   \\

        \bottomrule
      \end{tabular}
      }
      \vspace{-0.7cm}
    \end{table}

    For our second type of experiment, we perform action recognition for unseen types of animals. Specifically, we  pick video clips of six action classes (\ie, moving, eating, attending, swimming, sensing, and keeping still) that are widely shared by various types of animals, to train and test our CARe model for action recognition of unseen new types of animals. We select four animal types (\ie, birds, fishes, frogs, snakes) for training, and another five animal types (\ie, lizards, primates, spiders, orthopteran insects, water fowls) for testing. In this experiment, we use the video clips, each containing only one action instance. As such, we use accuracy as the evaluation metric for this experiment. We conduct ablation studies, as well as compare our method with two recent domain generalization methods \cite{li2019episodic, wang2020heterogeneous}, and all experiments are conducted by using I3D as the backbone. For the ablation study, we first construct `CARe without specific feature', and `CARe without general feature' (\ie, only using approximated specific feature). We also construct `CARe without spatially-aware weighting' where similarity weights $\textbf{w}$ are implemented as a set of scalars, while $\textbf{w}$ are a set of tensors in our full CARe model. Results in \Cref{table:ar_moe} show that our full model (CARe) achieves the best results, demonstrating the effectiveness of taking advantage of the general feature and also spatially-aware approximated specific features for recognizing actions of unseen types of animals.

    \begin{table}[t]
      \centering
      \caption{Results of action recognition of unseen animals}
      \label{table:ar_moe}
      \vspace{-0.3cm}
      \scalebox{0.75}{
      \begin{tabular}{cc}
        \toprule
        Method & Accuracy $(\%)$ \\
        \midrule
        Episodic-DG \cite{li2019episodic}  & 34.0  \\
        Mixup \cite{wang2020heterogeneous} & 36.2  \\
        \midrule
    
        CARe without specific feature & 27.3 \\
        CARe without general feature  & 38.2 \\
        CARe without spatially-aware weighting  & 37.1 \\
        CARe (Our full model) & 39.7 \\
        \bottomrule
      \end{tabular}
      }
      \vspace{-0.4cm}
      
    \end{table}

    \subsection{Video Grounding Results}
    Following \cite{gao2017tall, zhang2020span, mun2020LGI}, we use both mean Intersection over Union (IoU), and ``Recall@$n$, IoU=$\mu$'' as our evaluation metrics. Mean IoU means the average IoU of all test samples. As for ``Recall@$n$, IoU=$\mu$'', we first compute the IoU of the results with ground truth (\ie, length between the start and end time). The returned result is considered correct if IoU is larger than the threshold $\mu$. Hence, ``Recall@$n$, IoU=$\mu$'' refers to the percentage of language queries in the test set with at least one correct result among the top-$n$ returned results. We use $n$=1 and $\mu\in \{0.1,0.3,0.5,0.7\}$, and report the results in \Cref{table:vg}.

    \begin{table}[h]
      \centering
      \vspace{-0.3cm}
      \caption{Results of video grounding}
      \vspace{-0.36cm}
      \label{table:vg}
      \scalebox{0.7}{
      \begin{tabular}{cccccc}
        \toprule
            &   \multicolumn{4}{c}{Recall@$1$} & mean IoU \\
        \cmidrule(lr){2-5}
        Method                     & IoU=0.1 & IoU=0.3 & IoU=0.5 & IoU=0.7 &  \\
        \midrule
        LGI\cite{mun2020LGI}       & 50.84 & 33.51   & 19.74   & 8.94   & 22.90 \\
        VSLNet\cite{zhang2020span} & 53.59 & 33.74   & 20.83   & 12.22   & 25.02\\
        \bottomrule
      \end{tabular}
      }
      \vspace{-0.4cm}
    \end{table}

    \subsection{Pose Estimation Results}
    We define three protocols to identify potential challenges of pose estimation that arise due to intra and inter-animal class variations.

    \textbf{Protocol 1:} The whole dataset with all animal species is split to the training set (80\% of samples) and test set (20\% of samples). This means all species are contained in both the train and test sets, for evaluating the efficacy of models when estimating pose of all animals.
    
    \textbf{Protocol 2:} We adopt the leave-$k$-out setting by selecting $k=12$ animal sub-classes (\eg, felines, turtles, etc.) that are reserved and used only for testing, while the other animal sub-classes are used for training. The selected sub-classes come from the five major classes. Therefore, the sub-classes in the test set do not appear in the training set. This evaluates the generalization ability of the model for pose estimation of unseen animal classes.
    
    \textbf{Protocol 3:} We group all the samples in our dataset to five major classes (\ie, mammals, amphibians, reptiles, birds, and fishes). And for each major class, we define its own training set (80\% samples) and test set (20\% samples). This is to evaluate the pose estimation performance of each major class of animals. 
    
    Following \cite{yang2012articulated}, we adopt Percentage of Correct Keypoints (PCK) as our evaluation metric. PCK@$\alpha$ measures the accuracy of localizing body joints within the distance threshold $\alpha$. The predicted keypoint is considered to be correct if it falls within the distance threshold calculated by \begin{math} \alpha \times max(height, width) \end{math} of the animal bounding box. Here we set $\alpha$ to 0.05. As shown in \Cref{table:pe}, we evaluate two state-of-the-art pose estimation methods \cite{SunXLW19,Zhang_2020_CVPR} on our dataset.
    In Protocol 3, the results for reptiles and amphibians are lower than other animals, possibly due to the challenges of estimating their keypoints, as they come in various forms of shapes and textures with varying positions of their joints within its class, and they camouflage well into their environments. We observe that the result of Protocol 2 is comparatively lower than the rest, illustrating the challenges in generalizing the pose to unseen new animal classes. This shall bring opportunities for the community to develop novel methods to better learn and generalize pose information across various animal classes.
    
    \begin{table}[t]
        \centering
        \caption{Results of pose estimation}
        \label{table:pe}
        \vspace{-0.3cm}
        \scalebox{0.7}{
    \begin{tabular}{cccc} 
    \toprule
                       &                & \multicolumn{2}{c}{PCK@0.05}  \\ 
    \cline{3-4}
    Protocol           & Description    & HRNet \cite{SunXLW19} & HRNet-DARK \cite{Zhang_2020_CVPR}           \\ 
    \cline{1-4}
    Protocol 1                  & All  & 66.06 & 66.57              \\
    \midrule
    Protocol 2                  & Leave-$k$-out    & 39.30 & 40.28                 \\
    \midrule
    \multirow{5}{*}{Protocol 3} & Mammals        & 61.59 & 62.50               \\
                       & Amphibians     & 56.74 & 57.85                 \\
                       & Reptiles       & 56.06 & 57.06               \\
                       & Birds          & 77.35 & 77.41                \\
                       & Fishes         & 68.25   & 69.96                   \\

    \bottomrule
    \end{tabular}
    }
    \vspace{-0.5cm}
    \end{table}

    \section{Conclusion}
    
    We introduce a challenging animal behavioral analysis dataset, comprising over 850 species of animals in the animal kingdom for video grounding, action recognition, and pose estimation. We also introduce a Collaborative Action Recognition (CARe) model with improved ability to recognize actions of unseen new animal types. The variations within and between classes of animals for both action recognition and pose estimation illustrate the advantage of our dataset being diverse in nature, and yet at the same time highlight the challenges and reinforce the compelling need for further research in the areas of video grounding, animal action recognition and pose estimation. We expect that our work will inspire further works in animal behavioral analysis, and understanding animal behaviors can make our world a better place for both wildlife and humans.

    \setlength{\parindent}{0em}\setlength{\baselineskip}{0.1em}{
    {\scriptsize{
    \textbf{Acknowledgements.}
    We would like to thank Ang Yu Jie, Ann Mary Alen, Cheong Kah Yen Kelly, Foo Lin Geng, Gong Jia, Heng Jia Ming, Javier Heng Tze Jian, Javin Eng Hee Pin, Jignesh Sanjay Motwani, Li Meixuan, Li Tianjiao, Liang Junyi, Loy Xing Jun, Nicholas Gandhi Peradidjaya, Song Xulin, Tian Shengjing, Wang Yanbao, Xiang Siqi, and Xu Li for working on the annotations and conducting the quality checks for video grounding, action recognition and pose estimation. This project is supported by AI Singapore (AISG-100E-2020-065), National Research Foundation Singapore, and SUTD Startup Research Grant.}}}

    {\small
    \bibliographystyle{ieee_fullname}
    \bibliography{main.bib}
    }
    \end{document}

\title{Animal Kingdom: A Large and Diverse Dataset for Animal Behavior Understanding (Supplementary Materials)}

\author{
Xun Long Ng{\textsuperscript{‡}}\quad 
Kian Eng Ong{\textsuperscript{‡}}\quad
Qichen Zheng{\textsuperscript{‡}}\quad
Yun Ni{\textsuperscript{‡}}\quad 
Si Yong Yeo \quad
Jun Liu\thanks{Corresponding author.} \\
\normalsize Information Systems Technology and Design, 
Singapore University of Technology and Design, Singapore 
\\
{\tt\scriptsize \{xunlong\_ng, kianeng\_ong\}@mymail.sutd.edu.sg} ~~~~~~~
{\tt\scriptsize \{qichen\_zheng, ni\_yun, siyong\_yeo, jun\_liu\}@sutd.edu.sg}
}

\maketitle

\def\thefootnote{{\textsuperscript{‡}}}\footnotetext{Equal contribution to this work.}

\section{Details of Data Collection and Verification} 
\label{dataset}
The team consisted of 23 members, including biology experts with knowledge of biodiversity.

We manually identify and provide framewise annotations of both animal and action descriptions for over 50 hours of videos that were collected from YouTube videos. We process them into a total of 30,100 video clips, each ranging from 1 second to 117 seconds (average 6 seconds). The title, description or captions in the videos contains the name of the animals. We manually check and tally the animals in the video clips with the animal names provided. As for the actions, we identify them using a defined set of ethological terms \cite{mouse_ethogram, stafford2011inferential, rose2021conducting, NC3RS_ethogram}. In order to minimize discrepancies in our annotations, we first identify the commonly used terms that describe the same type of action in different classes of animals (\eg, grooming in insects, but preening in birds refers to the same act of self-maintenance), and redefine ambiguous terms that describe vastly different movements (\eg, snake gliding on land versus eagle gliding in the sky). At the same time, we also include descriptions of the scenes, and note their start and end times for video grounding task. We conduct a total of three rounds of quality checks through various permutations of cross-checks by different individuals to verify the correct start and end time of the video clips for video grounding task, and harmonize the nomenclature of actions for action recognition task. 

\par

As for pose estimation task, we label a total of 33K images that are extracted from the video clips, using Label Studio \cite{LabelStudio}. Annotators are given instructions with reference images on how to label each class of animals. As the animal footages are captured in the wild, the complex backgrounds with various illumination and weather conditions make the annotation challenging. Thus, three rounds of quality checks are performed to ensure that the keypoints are correctly labelled.

\newpage
\section{Diverse Range of Animals}
\label{animals}
\begin{figure}[h]
    \centering
\includegraphics[width=0.48\textwidth]{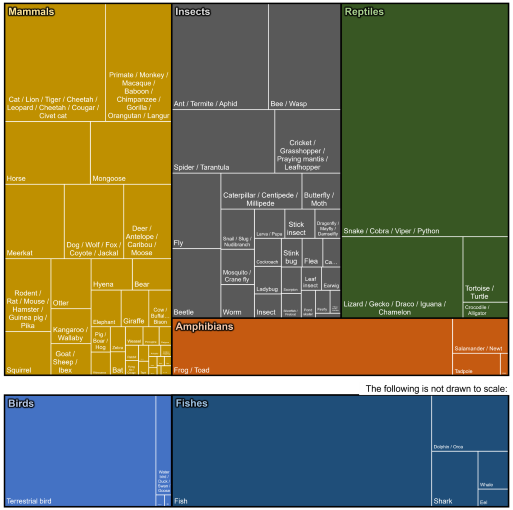}
\caption{Distribution of clips of over 850 species of animals in 6 major animal classes classified based on their appearance, number of limbs and how they move, and further divided into sub-classes.}
    \label{fig:dist_animal}
\end{figure}

\begin{figure}[h]
    \centering
\includegraphics[width=0.49\textwidth]{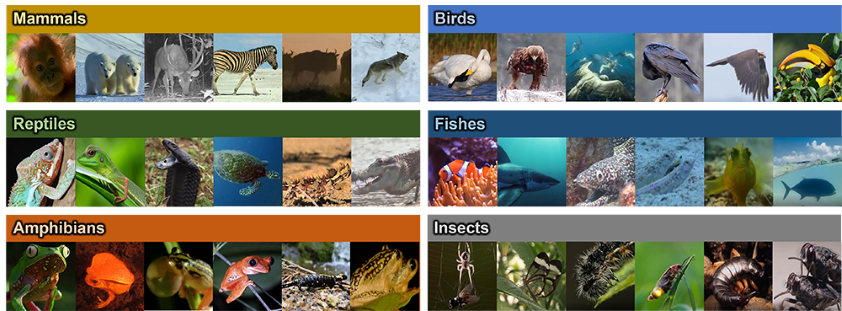}
    \caption{Examples of animals in 6 main animal classes}
    \label{fig:eg_animal}

\end{figure}

Our dataset contains over 850 species of animals. We group them into 6 major animal classes, and further divide them into sub-classes. 
The distribution of animals in our dataset is shown in \cref{fig:dist_animal} and examples are shown in \cref{fig:eg_animal}.

For action recognition, we label the actions of animals in 6 major classes (\ie, mammals, reptiles, amphibians, birds, fishes, insects). For pose estimation, we label the poses of animals in 5 major classes (\ie, mammals, reptiles, amphibians, birds, fishes). Because different animals can have vastly different anatomical structures across classes (\eg, insects vs mammals), which poses a great challenge for pose annotation, we thus label these five animal classes which share similarities in their anatomical structures.

\newpage

\section{Diverse Range of Actions} 
\label{actions}
Our dataset contains a diverse range of 140 actions. The collection of actions and behaviors encompasses: \par
(1) movement, \eg, swimming, running, flying, \par
(2) transport, \eg, carrying in mouth,  \par
(3) feeding, \eg, eating, biting, drinking, \par
(4) sensing, \eg, exploring, attending, \par
(5) resting, \eg, sleeping, \par 
(6) maintenance, \eg, grooming, washing, \par
(7) communication, \eg, chirping,  \par
(8) aggressive, \eg, attacking, spitting venom, \par
(9) defensive, \eg, retreating, displaying defensive pose, \par
(10) social, \eg, playing, \par
(11) affection, \eg, hugging, \par
(12) sexual, \eg, sexual display, copulation, \par
(13) life events, \eg, giving birth, laying eggs, hatching, \par
(14) other general actions, \eg, panting, flapping ears. 

Hence, our dataset covers a broad range of actions seen in nature.

\section{More Examples} 
\begin{figure}[h]
  \centering 
  \includegraphics[width=0.48\textwidth]{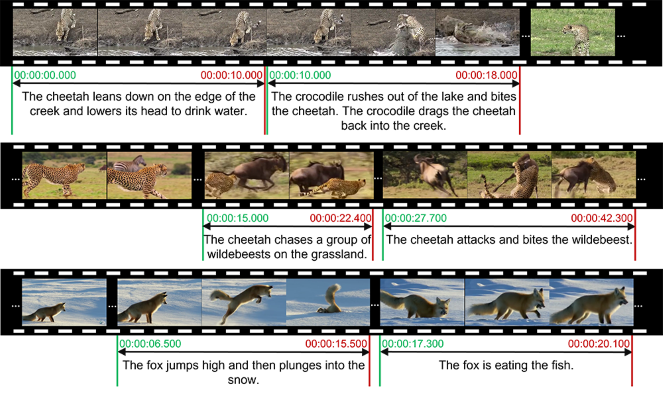}
    \caption{Samples of the video grounding task. Given the language description, we need to detect the corresponding time sequence.
    }
    \label{fig:eg_vg_suppl}
    \vspace{-0.3cm}
\end{figure}

\begin{figure}[h]
  \centering 
  \includegraphics[width=0.48\textwidth]{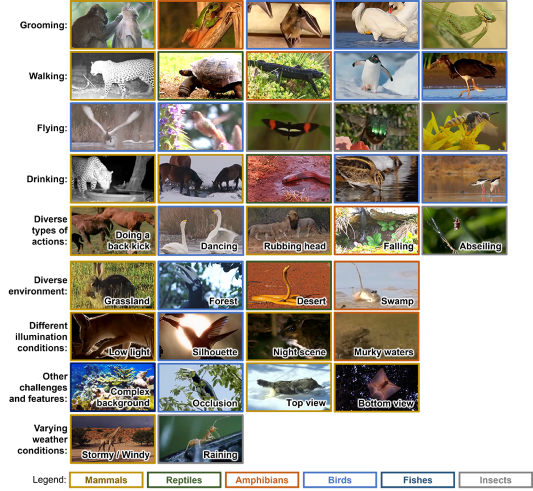}
    \caption{More examples of actions. Rows 1 to 4 show how the same set of actions differ across various animal classes. Row 5 shows examples of various actions in our dataset. Rows 6 to 9 show sample features of our dataset, ranging from diverse environments to varying illumination and weather conditions.
    }
    \label{fig:eg_ar_suppl}
    \vspace{-0.1cm}
\end{figure}

\begin{figure}[h]
  \centering 
  \includegraphics[width=0.48\textwidth]{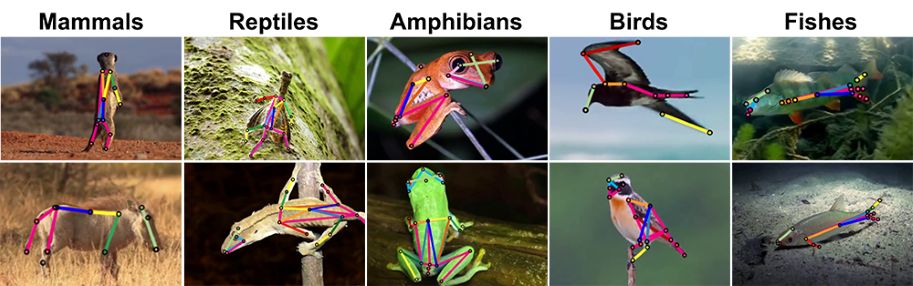}
    \caption{Examples of animal poses in our dataset}
    \label{fig:eg_pe_suppl}
    \vspace{-0.2cm}
\end{figure}

\begin{figure}[h]
  \centering 
  \includegraphics[width=0.39\textwidth] {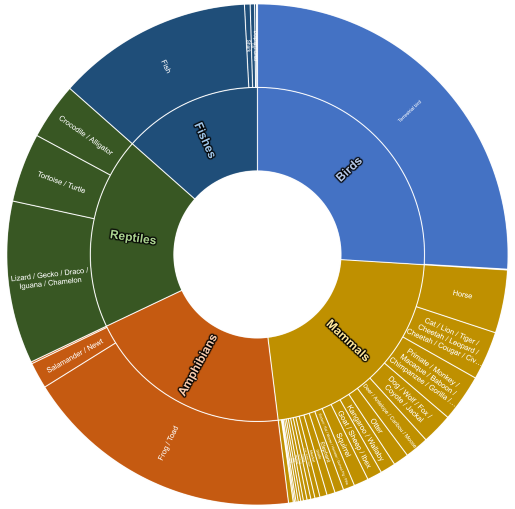}
    \caption{Distribution of the 33K animal pose annotations in the 5 major animal classes}
    \label{fig:dist_pe}
    \vspace{-0.2cm}
\end{figure}

Here, we provide more examples of our video grounding (\cref{fig:eg_vg_suppl}), action recognition (\cref{fig:eg_ar_suppl}), and pose estimation (\cref{fig:eg_pe_suppl}) tasks. The distribution of animals with its pose annotated is illustrated in \cref{fig:dist_pe}.

\newpage

\section{Details of the Proposed CARe Model} 
\begin{figure}[ht]
  \centering 
    \includegraphics[width=0.98\columnwidth] {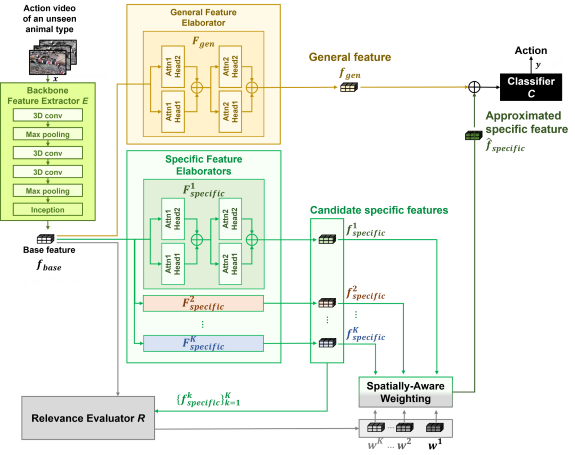}
    \caption{Architecture of our Collaborative Action Recognition (CARe) model.} 
    \label{fig:arch_supp}
    \vspace{-0.3cm}
\end{figure}

In this section, we present the implementation details of our CARe model. We discuss in detail the architecture of the model (\cref{fig:arch_supp}) and how it is trained (\cref{CARe_algo}).  

The backbone feature extractor $E$, the feature elaborators $F_{gen}$ and $\{F_{specific}^k\}_{k=1}^K$, and the classifier $C$ form the basic version of our model. Below we present the design of each of them. 

For the backbone feature extractor $E$, we adopt the early part of the I3D architecture. It includes three 3D convolutional layers $E\_{conv1}$, $E\_{conv2}$, and $E\_{conv3}$, two maxpooling layers $E\_{maxpool1}$ and $E\_{maxpool2}$, and an inception submodule of the I3D architecture $E\_{incept}$. Given the input video $x \in \mathbb{R} ^ {ch \times t \times h \times w}$ (whereby the number of channels $ch$, number of frames $t$, height $h$, width $w$ equals 3, 16, 180, and 320 respectively), it will be transformed by $E\_{conv1}$, $E\_{maxpool1}$, $E\_{conv2}$, $E\_{conv3}$, $E\_{maxpool2}$, and $E\_{incept}$ sequentially to obtain the base feature $f_{base} \in \mathbb{R}^{ch' \times t' \times h' \times w'}$, with $ch'=256$, $t'=8$, $h'=23$, and $w'=40$. 

Regarding the subsequent feature elaborators $\{F_{specific}^k\}_{k=1}^K$ and $F_{gen}$, each of them consists of two lightweight two-head self-attention layers \cite{girdhar2019video}. Given the base feature, the specific features $\{f_{specific}^k\}_{k=1}^{K}$ and general feature $f_{gen}$ computed by the feature elaborators have the size of ${ch'' \times h'' \times w''}$, where $ch''$, $h''$, and $w''$ equal 4, 12, and 20.  

The final classifier $C$ consists of one linear layer $C\_linear$ that maps the flattened and concatenated general and specific features to action likelihoods $y \in \mathbb{R}^M$ for the $M$ possible actions.   

Besides components of the basic version of the CARe model, the relevance evaluator $R$ is included to compute similarity scores for unseen animals. It consists of two layers, $R\_{conv1}$ and $R\_{conv2}$, which will be used to further transform the base feature, and $K$ layers $\{R\_linear^k\}_{k=1}^K$ to compute the set of similarity scores $\{w^k\}_{k=1}^K$. More specifically,
$f_{base}$ will be first transformed by $R\_{conv1}$ and $R\_{conv2}$. To obtain the similarity score for the $k$-th specific feature elaborator $w^k$, we flatten and concatenate the transformed base feature and the specific feature $f_{specific}^k$, which will be fed into its respective layer $R\_{linear^k}$ to compute $w^k$.

\textbf{Training and testing.} Since the base feature extractor follows the I3D architecture, we initialize its parameters with the pre-trained I3D. The parameters of the base feature extractor will be optimized with an initial learning rate of 0.001. Other components of our CARe model will be randomly initialized and given a higher initial learning rate of 0.01. SGD optimizers are used to update all components. A summary of the training scheme can be found in Algorithm \ref{CARe_algo}.

%-----------------------Algorithm------------------------------
\begin{algorithm}[tbh]

\footnotesize
\LinesNumbered
\caption{\textbf{Training Procedures of CARe Model}}
\label{CARe_algo}
\SetKwData{Left}{left}\SetKwData{This}{this}\SetKwData{Up}{up}
  \SetKwFunction{Union}{Union}\SetKwFunction{FindCompress}{FindCompress}
  \SetKwInOut{Input}{Input}\SetKwInOut{Output}{Output}
\Input{Data of $K$ animal types 
$\{D_k\}_{k=1}^K$ \\
Learning rates  $\alpha$, $\beta$ \\ Hyper-parameter $\lambda$}
\Output{Backbone feature extractor $E$ \\
General feature elaborator $F_{gen}$ \\
Specific feature elaborators $\{F^k_{specific}\}_{k=1}^K$ \\
Classifier $C$ \\
Relevance evaluator $R$}
\While{not converged}{
            \textbf{1. Update} $E$, $F_{gen}$, $\{F^k_{specific}\}_{k=1}^K$, $C$:~\\
            Calculate the cross-entropy losses $\{{\ell^k}\}_{k=1}^K$ for $\{{D^k}\}_{k=1}^K$ using $E$, $F_{gen}$, their respective $\{F^k_{specific}\}_{k=1}^K$, and $C$\;
            Update the parameters for $E$, $F_{gen}$, 
            $\{F^k_{specific}\}_{k=1}^K$ and $C$\;
            
             \textbf{2. Update $R$:} ~\\
                Sample 1 type of animal as the meta-test 
                $D_{mtest}$ and the other $K$-1 types of animal
                as meta-train $D_{mtrain}$\;
                \textbf {2.1 Meta-train:}~\\
                    Calculate cross-entropy losses ${{\ell}_{mtrain}}$ for $D_{mtrain}$\;
                    Update the parameters for $R$ using: ~\\
                     ${\phi}'\gets\phi-\alpha\triangledown_{\phi}{\ell}_{mtrain}(\phi)$\;
                \textbf {2.2 Meta-test:}~\\
                    Calculate the cross-entropy loss ${{\ell}_{mtest}}$
                    for $D_{mtest}$\;
                \textbf {2.3 Meta update:}~\\
                    Update the parameters $\phi$ of $R$ using: ~\\
             $\phi\gets\phi-\beta((1-\lambda)\triangledown_{\phi}{\ell}_{mtrain}(\phi)+\lambda\triangledown_{\phi}{\ell}_{mtest}({\phi'}))$\;
        }

\end{algorithm}\DecMargin{1em}
In our experiments, 16 frames from each video sequence are randomly selected and used as the input. The model was trained for 40 epochs with a 2-GPU implementation. The initial learning rates were used for the first 30 epochs. The reduced learning rates that are 10\% of the initial ones were used to train the model for 10 additional epochs.  

\newpage
{\small
\bibliographystyle{ieee_fullname}
\bibliography{supp.bib}
}